%% file: main.tex
\documentclass[11pt]{article}

\usepackage[T1]{fontenc}
\usepackage{lmodern}
\usepackage[margin=1in]{geometry}
\usepackage{microtype}

\usepackage{amsmath}
\usepackage{amssymb}      
\usepackage{booktabs}
\usepackage{tabularx}
\usepackage{array}
\usepackage{xcolor}
\usepackage{colortbl}     
\usepackage{graphicx}

\usepackage[colorlinks=true,linkcolor=blue,citecolor=blue,
            urlcolor=blue]{hyperref}

\usepackage[backend=biber,
            style=numeric,
            sorting=none,
            maxbibnames=6,
            minbibnames=3]{biblatex}
\definecolor{aitellbg}{RGB}{255,230,230}
\definecolor{goodbg}{RGB}{220,245,220}
\definecolor{midbg}{RGB}{255,248,220}
\definecolor{imsrow}{RGB}{232,240,254}
\newcommand{\aitell}[1]{\colorbox{aitellbg}{\strut #1}}
\newcommand{\changed}[1]{\colorbox{goodbg}{\strut #1}}
\newcommand{\partialfix}[1]{\colorbox{midbg}{\strut #1}}

\title{Small Is Enough: Per-User Style Rewriting of\\
  AI-Edited Text via LoRA Adapters%
  \thanks{Accepted at the 2026 9th International Conference on
  Computational Intelligence and Intelligent Systems (CIIS 2026),
  October 26--28, 2026, Osaka, Japan. This is the author's preprint
  version; the version of record will appear in the ACM International
  Conference Proceedings Series.}}

\author{Antorweep Chakravorty\\
  University of Stavanger, Stavanger, Norway\\
  \texttt{antorweep.chakravorty@uis.no}\\
  ORCID: 0000-0002-3576-2406}

\date{}

\begin{document}

\maketitle

\begin{abstract}
\input{sections/00_abstract}
\end{abstract}


\bigskip

\input{sections/01_introduction}

\input{sections/02_related_work}
\input{sections/03_method}
\input{sections/04_experimental_setup}
\input{sections/05_results}
\input{sections/06_limitations}
\input{sections/07_conclusion}

\printbibliography

\appendix
\input{sections/08_qualitative_appendix}

\end{document}

%% file: sections/00_abstract.tex
InMyStyle is a privacy-first, single-user system that adapts small language
models to rewrite AI-edited text towards an individual user's writing style
without an instruction prompt at inference. Given a user's documents, it uses
multiple local helper LLMs to construct paired training examples and
fine-tunes LoRA adapters on Qwen2.5 models ranging from 0.5B to 7B
parameters. Length-aware generation budgets and automatic chunking support
inputs of different lengths. We report a single-user case study: 219 evaluation
pairs derived from 73 paragraphs of one author's scientific writing, with all
adapters trained using the same rank-8, three-epoch recipe. The automatic
composite score (0-1 scale) plateaus across model sizes under both greedy
and sampled decoding ($Q=0.689$-$0.695$, with overlapping confidence intervals). In this setting, small models are sufficient for the
measured rewriting task, and model size mainly determines efficiency trade-offs
rather than a stable quality ranking. The gains favor content-preserving
naturalization more than recovery of personal style, with authorship probabilities staying near the classifier's
decision boundary (0.51--0.53) and stylometric improvement being
near zero. As a secondary evaluation, 400 ratings from five LLM judges
give InMyStyle outputs a mean perceived AI-ness score over 20\% lower than
their helper-generated inputs, with scores decreasing with model size in this
sample. The study does not establish generalization across users. 

%% file: sections/01_introduction.tex
\section{Introduction}
AI-assisted editing can create a trade-off where users want their prose to be
more clear, but do not want to sacrifice their idiosyncratic wording, sentence
structure, and framing. This creates a personalization problem: can an AI-edited paragraph be rewritten in the direction of a specific user's writing
style while preserving its content? Our goal is not to evade AI-text detectors,
but personalized rewriting. 

InMyStyle\footnote{Project source: \url{https://github.com/achak1987/inmystyle}} is a privacy-first, single-user system that learns this
transformation from the user's own documents. Multiple local
helper LLMs convert the original paragraphs into ``AI-shadow''
variants, which are then paired with the originals to train the
LoRA adapters~\cite{hu2021lora} on four compact Qwen2.5 models~\cite{qwenQwen25TechnicalReport2025} with parameter
sizes of 0.5B, 1.5B, 3B and 7B. Response-only loss restricts
the optimization to the user-written target tokens. Three requirements shaped the design: no explicit style
instruction at inference time, no external processing of personal
documents, and a separate adapter for each user rather than a generic ``humanization'' model.

Our main question is how much base-model capacity the measured rewriting task requires. We
hypothesize that recurring preferences in wording, sentence structure and framing can be captured
through low-rank updates without the need to train the complete model. We evaluate
the models on 219 held-out pairs derived from 73 scientific-paper paragraphs under greedy and
sampled decoding. All model and decoding combinations achieve automatic composite scores 
between 0.689 and 0.695 (scale 0--1), 
forming an observed plateau across model sizes. This suggests that the measured
task can be handled by compact models with the given dataset and training recipe with smaller sizes being better suited  when compute is limited.

As a secondary evaluation, five LLM judges provide 400
perceived-AI ratings. Pooled across judges, InMyStyle rewrites
receive a 22\% lower mean perceived-AI score than their helper generated inputs. The ratings vary substantially across judges,
with the size-pooled means within InMyStyle decreasing as model
size increases. This secondary result is descriptive and does not overturn the automatic plateau.

We distinguish two goals that are easily conflated. \emph{Naturalization} removes
surface markers of AI editing while preserving content. \emph{Personal-style
recovery} moves the text measurably toward the target user's own writing,
as captured by authorship and stylometric measures. Our results provide better evidence for the former than the latter. All experiments correspond to a single-user case study comprising of their own
corpus, Qwen2.5 model family, and a LoRA rank 8 based three epochs training recipe. We thus report the size plateau as an observation
for this particular setting, and do not propose a general model-size law or
generalization across users in this work.

Our contributions are: (1) a local pipeline for constructing multi-family AI-shadow to user-original training pairs with paragraph-level split isolation and length-aware rewriting; (2) a controlled
model-size and decoding study; and
(3) an LLM-as-a-judge analysis of perceived AI-ness.

%% file: sections/02_related_work.tex
\section{Related Work}
Text style transfer aims at changing the
stylistic properties while keeping the underlying content~\cite{jin2022deep}. Authorship transfer is particularly challenging as the target is a
user's recurring choices rather than a predefined attribute. StyleRemix uses pretrained LoRA modules to perturb interpretable
stylistic dimensions for authorship obfuscation~\cite{fisher2024styleremix}. TinyStyler combines an 800M-parameter model with
authorship embeddings for few-shot transfer to a target style~\cite{horvitz2024tinystyler}, while AuthorMix composes author-specific
LoRA adapters layer by layer to support low-resource target authors~\cite{thillainathan2026authormix}. InMyStyle differs as it constructs
paired AI-shadow to user-original examples from the user's documents and trains a dedicated rewriting adapter. At inference, it requires neither target examples nor an explicit style instruction.

LoRA enables compact task- or user-specific adaptation by
freezing the base model and learning low-rank updates~\cite{hu2021lora}. This makes it practical to store a separate adapter
for each user. OPPU similarly represents user-specific behavior
through individual PEFT modules~\cite{tan2024democratizing}.
Panza is closest to our privacy setting: it provides a fully
local personalized email assistant using fine-tuning, reverse
instructions, and retrieval-augmented generation~\cite{nicolicioiu2024panza}. InMyStyle instead targets paragraph rewriting with
a single local adapter and no retrieval component at inference. We do not introduce a new PEFT algorithm; the contribution
is a private paired-data pipeline and a controlled comparison
of LoRA adaptation.

Stylistic transfer is inherently multi-faceted, as stronger
stylistic change can reduce semantic fidelity or fluency. Automatic metrics may also disagree with human judgments,
motivating complementary evaluation with language models
or people~\cite{ostheimer2024evaluation}. InMyStyle reports
a composite of authorship, content, target similarity, AI-tell
reduction, and stylometric measures, and includes a separate
LLM-as-a-judge study. Work on AI-text detection provides
additional context for evaluating rewritten machine text:
RAID finds substantial degradation across unseen generators,
decoding strategies, and adversarial transformations~\cite{dugan2024raid}, while controlled paraphrasing can sharply
reduce detector accuracy without large semantic changes~\cite{krishna2023paraphrasing}. InMyStyle does not train against
or evaluate detector evasion; this work provides context for
examining how rewriting affects the perceived and measurable characteristics of AI-edited prose.

%% file: sections/03_method.tex
\section{Method}
\label{sec:method}

\subsection{Task Formulation and Design Constraints}
InMyStyle learns a separate paragraph-rewriting function per user. Let
$\mathcal{Y}_u = \{y_i\}_{i=1}^{N}$ be the paragraphs extracted from user $u$'s
documents. For each original paragraph $y_i$, a helper model $h_k$ produces
an AI-shadow paragraph that conveys the same content, but in machine generated prose. After optional corruption, this paragraph is fed as input
$x_{ik}$ to the paragraph-rewriting model, leading to the paired dataset
\begin{equation}
    \mathcal{D}_u = \{(x_{ik},y_i) : i=1,\ldots,N;\ k=1,\ldots,K\},
\end{equation}
where $N$ is the number of retained user paragraphs, $K$ is the number
of helper models, and $i$ and $k$ index source paragraphs and helper
models, respectively. A frozen language model with user-specific adapter
parameters $\phi_u$ is trained to map $x_{ik}$ back to $y_i$. The objective is not to reconstruct the
exact source paragraph for arbitrary inputs, but to learn recurring choices in
the user's prose while retaining the input's meaning.

Three constraints determine the system design. First, inference must not
depend on an explicit instruction such as ``rewrite this in my style'' or on
retrieved style demonstrations. Second, personal documents and all derived
training text must remain on the user's machine. Third, personalization
must be represented by a separate adapter for each user, rather than by a
shared multi-user model. The model's tokenizer still supplies its ordinary
chat formatting, but the inference input contains only the paragraph to
rewrite without any explicit system message or natural-language rewriting instruction.

\subsection{Corpus Preparation}
\label{sec:corpus-preparation}
The source corpus comprises documents written by the target user. InMyStyle
extracts the page text from the source corpus PDF files and passes it through two
stages of filtering before creating the training pairs. The first stage performs
deterministic mechanical cleanup while ingesting the pages. It normalizes Unicode, fixes common mojibake (garbled text), rejoins words broken by line-end
hyphenation, removes inline citation markers, and discards page numbers, captions,
URLs, spaced-letter headings, and repeated headers or footers. The cleaned pages
are stored in a local SQLite database and split into paragraph candidates.

The second stage uses paragraph-level content heuristics
outside the database. It filters out candidates that resemble
flattened tables or lists, URL- or email-heavy fragments, affiliation blocks, math-dense passages, and cross-page
fragments. We purposefully separate
these two stages: the mechanical cleaning provides a stable
representation of the source documents, while the heuristic
thresholds can be revised and re-run without re-extracting the
PDFs. Rather than fixing highly corrupted passages, the
pipeline simply ignores them, allowing adapter capacity to
focus on continuous prose.

\subsection{Multi-Family AI-Shadow Construction}
\label{sec:shadow-construction}
Supervised style adaptation requires paired machine-style
inputs and user-written targets. However, naturally occurring
aligned pairs are rare. InMyStyle creates the input side by
automatically paraphrasing the user's paragraph while keeping
it unchanged as the target. Three locally executed helper
models from different model families: Qwen2.5-3B, Llama-3.2-3B, and Phi-3-mini~\cite{qwenQwen25TechnicalReport2025,
grattafioriLlama3Herd2024, abdinPhi3TechnicalReport2024};
each paraphrases every retained paragraph once. The common
helper instruction is to provide a meaning-preserving rewrite
and suppress explanations, markup, and preambles. The outputs
are subsequently cleaned to remove any remaining introductions, metadata, quotation fences, and repeated-token
failures. The generation budget for each helper rewrite scales with the length of the source paragraph rather
than using the same fixed allowance for every input.

The use of multiple helper families increases the variety of
machine phrasings seen during training. A single helper could otherwise bias
the learned transformation to specific lexical and structural patterns of that
model. We further use two stochastic corruption operators. With probability
$0.5$, an AI-tell operator replaces common words with more stereotypically
machine-like alternatives, such as \emph{use} with \emph{utilize}, or inserts
formulaic framing expressions. With probability $0.3$, a light-noise operator
drops a small fraction of words and occasionally transposes single characters. Thus,
\begin{equation}
    x_{ik} = C\!\left(h_k(y_i)\right),
\end{equation}
where $C$ denotes the composition of the optional AI-tell and noise operators.
The paragraph identifier, helper identity, applied corruptions and input and
target sentences are retained as provenance for every pair.

\subsection{Leakage-Resistant Dataset Construction}
\label{sec:grouped-split}
Each original paragraph produces several rows with the same target. A
random row-level split would cause target text leakage: one helper variant of
$y_i$ could end up in training while another variant of the same $y_i$ ends up
in evaluation. InMyStyle instead partitions the set of source paragraph
identifiers and assigns every variant associated with one identifier to the
same partition. Let $\mathcal{D}_{\mathrm{train}}$ and
$\mathcal{D}_{\mathrm{eval}}$ denote the resulting training and evaluation
partitions, respectively. Consequently,
\begin{equation}
    \{i:(x_{ik},y_i)\in\mathcal{D}_{\mathrm{train}}\}
    \cap
    \{i:(x_{ik},y_i)\in\mathcal{D}_{\mathrm{eval}}\}
    = \varnothing.
\end{equation}
This grouped split evaluates transfer to unseen source paragraphs rather than
recall of a target observed through another helper model. Each retained pair is
stored as a two-turn conversation: $x_{ik}$ occupies the user turn and $y_i$
the assistant turn. No system turn is added.

\subsection{User-Specific LoRA Adaptation}
\label{sec:lora-training}
For a pretrained model with parameters $\theta$, InMyStyle freezes the base weights
and inserts LoRA updates into its linear layers~\cite{hu2021lora}. For a weight
matrix $W$, the adapted transformation is
\begin{equation}
    W' = W + \frac{\alpha}{r}BA,
\end{equation}
where $A$ and $B$ are trainable low-rank matrices, $r$ is the adapter rank, and $\alpha$
controls the update scale. Only these adapter parameters are optimized and
stored for the user. The same training interface supports conventional LoRA
and 4-bit QLoRA~\cite{dettmersQLoRAEfficientFinetuning2023}, allowing the base
weights to remain frozen while reducing memory use on supported hardware.

The model is trained using a response-only loss. The input
tokens are included as conditioning context, but are masked from the
objective; the cross-entropy is computed only over the user-written
assistant response:
\begin{equation}
    \mathcal{L}(\phi_u)
    = -\sum_{(x,y)\in\mathcal{D}_{\mathrm{train}}}
      \sum_{t=1}^{|y|}
      \log p_{\theta,\phi_u}\!\left(y_t\mid x,y_{<t}\right).
\end{equation}
Here, $x$ is the AI-shadow input, $y$ is the user-written target, $t$
indexes target tokens, $|y|$ is the target length and $y_{<t}$ are the
target tokens preceding position $t$. The term $p_{\theta,\phi_u}$ is the
next-token distribution defined by the frozen base parameters $\theta$
and the user-specific adapter parameters $\phi_u$.

The masking is crucial as the input is
deliberately machine-edited and could suffer from further corruption. Including
the input tokens in the objective would train the adapter to model the same
machine-style text as it is trying to transform. Hence, for each AI-shadow user-original pair, the loss is computed only over the user-written target tokens,
with the AI-shadow tokens serving merely as conditioning context. The model
sizes, adapter geometry, optimization settings and the hardware-specific
quantization configuration are described in the experimental setup.

\subsection{Prompt-Free Rewriting}
\label{sec:inference}
At inference time, the base model and the adapter for the selected
user are loaded. The input paragraph is fed directly to the tokenizer's
user turn and generation starts from the assistant turn. The system
does not provide any style label, rewriting instruction, target
example or retrieval context. Greedy decoding is used as a default
since it provides deterministic paragraph rewrites. Sampling is
an optional alternative that is evaluated separately.

Generation is guided by a length-aware token budget. For an input
segment with $w$ words, the budget is $1.5w+32$ new tokens, floored
to the budget for the configured minimum paragraph length of 30
words and capped at a hard limit of 600 tokens. This allows shorter
paragraphs less room for padding or repetition while enabling longer
inputs to expand when needed. Inputs with at most 500 words are
kept as a single segment. Longer inputs are greedily split at sentence
boundaries, each segment is independently rewritten and the outputs
are concatenated with blank lines between segments.

%% file: sections/04_experimental_setup.tex
\section{Experimental Setup}
\label{sec:experimental-setup}

\subsection{Objectives}
The experiments assess whether increasing the
base-model size yields a consistent advantage in automatic style-rewriting quality when the training data and adapter configuration is kept fixed. Moreover, we investigate whether stochastic decoding
outperforms deterministic greedy decoding. Finally, we use
five LLM judges as model-dependent proxies for perceived
AI-ness.

\subsection{Dataset}
\label{sec:experimental-corpus}
The main corpus includes around 36 scientific papers
published between 2012 and 2025 in peer-reviewed journals
and conference proceedings, authored or co-authored by the
target user. The papers are related to privacy-preserving data
analytics, distributed and cloud computing, blockchain-based
systems, energy informatics, and machine-learning applications.
After PDF extraction and the two-stage filtering process
explained in Section~\ref{sec:corpus-preparation}, 487 prose
paragraphs remained. The three helper models provided one
AI-shadow for each paragraph, which generated 1,461 paired
examples. We re-generated the variants with the length-aware
helper budgets described in Section~\ref{sec:shadow-construction} before
retraining and evaluating all four adapters.

We split the data by source paragraph identifier using an
evaluation fraction of $0.15$. The training partition included
414 source paragraphs and 1,242 pairs. The evaluation
partition included 73 source paragraphs and 219 pairs. All
three helper variants of a source paragraph remained in the
same partition. We excluded pairs whose combined input
and target length exceeded 8,192 characters before constructing the dataset. The final counts above reflect the records
that remained after this exclusion.

\subsection{Configuration}
\label{sec:experimental-models}
We trained adapters for the instruction-tuned Qwen2.5 models with 0.5B, 1.5B,
3B, and 7B parameters~\cite{qwenQwen25TechnicalReport2025}. Using checkpoints
from the same model family reduces cross-family differences in tokenization,
vocabulary, and model design, making model size the primary experimental
variable. The evaluated checkpoints span from 0.5B to 7B parameters, a fourteen-fold range. Every adapter used the same paired dataset and optimization recipe.
Table~\ref{tab:training-config} summarizes the shared configuration.

\begin{table}[t]
    \caption{Training configuration shared by all four model sizes.}
    \label{tab:training-config}
    \centering
    \small
    \begin{tabular}{ll}
        \toprule
        Setting & Value \\
        \midrule
        LoRA rank / scaling & $8$ / $16$ \\
        LoRA dropout & $0.05$ \\
        Target modules & All linear layers \\
        Epochs & $3$ \\
        Per-device batch size & $2$ \\
        Gradient accumulation & $8$ \\
        Effective batch size & $16$ \\
        Maximum sequence length & $2{,}048$ tokens \\
        Learning rate & $2\times10^{-4}$ \\
        LR schedule / warmup & Cosine / $0.03$ \\
        Weight decay & $0.01$ \\
        Random seed & $42$ \\
        Quantization & 4-bit NF4 QLoRA \\
        \bottomrule
    \end{tabular}
\end{table}

Training used response-only loss, length-grouped batches, and
gradient checkpointing. Following QLoRA, we reduced the model's
memory footprint by storing the base-model weights in 4-bit NormalFloat (NF4) format,
compressing their scaling values, and using bfloat16 for calculations
~\cite{dettmersQLoRAEfficientFinetuning2023}. All runs were performed
on one NVIDIA RTX PRO 5000 Blackwell Generation Laptop GPU. The
training environment used Python 3.12, PyTorch 2.13.0 with CUDA 13.2, Transformers 4.46.3, PEFT
0.13.2, TRL 0.12.1, and bitsandbytes 0.49.2.

\subsection{Decoding Conditions}
\label{sec:decoding-conditions}
Each adapter generated rewrites for all 219 evaluation pairs under two
decoding modes. Greedy decoding selects the highest-probability token at
each step. Sampled decoding uses temperature $0.7$ to control the sharpness
of the next-token distribution and nucleus sampling with top-$p=0.9$, which
limits sampling to the smallest token set whose cumulative probability
reaches $0.9$~\cite{holtzmanCuriousCaseNeural2019}. For both modes, 600
new tokens is a hard ceiling per segment rather than a fixed budget; the
actual per-segment budget follows the length-aware rule in
Section~\ref{sec:inference}. The same evaluation inputs and adapter weights
are used for both modes, leading to eight complete evaluations: four model
sizes times two decoding modes.

\subsection{Automatic Evaluation}
\label{sec:automatic-evaluation}
We used five complementary metrics to evaluate each rewrite.
\textbf{Authorship} ($A$) is the probability of the target-user class
predicted by a character-level TF-IDF logistic-regression classifier
following a standard linear approach to authorship attribution
~\cite{stamatatosSurveyModernAuthorship2009}. The classifier was
trained on the training partition only, using 414 original paragraphs
as positive examples and 414 randomly downsampled AI-shadow
inputs as negative examples. The classifier used character
$1-3$ grams within word boundaries and at most 10,000 features.

\textbf{Content similarity} ($C$) is the BERTScore F1 between the rewrite and
its AI-shadow input~\cite{zhangBERTScoreEvaluatingText2019}. It measures
whether the rewrite preserves the meaning of the text supplied at inference.
\textbf{Target similarity} ($T$) is the BERTScore F1 between the rewrite and the
corresponding user-written reference paragraph. It measures how closely the
rewrite approaches the paired target used during evaluation. Both metrics were
computed with \texttt{roberta-base}~\cite{liuRoBERTaRobustlyOptimized2019}.

\textbf{AI-tell reduction} ($R$) measures the relative decrease in
occurrences of the configured AI-associated phrases and the
substituted words:
\begin{equation}
    R = \operatorname{clip}_{[-2,1]}
        \left(
        \frac{n_{\mathrm{in}}-n_{\mathrm{out}}}
        {\max(n_{\mathrm{in}},1)}
        \right),
\end{equation}
here $n_{\mathrm{in}}$ and $n_{\mathrm{out}}$ are the respective tell counts in the AI-shadow input
and the rewrite. If the value is positive, it means that the rewrite was able to
remove the configured AI tells. If the value is negative, it means that the
rewrite has added additional tells.

\textbf{Stylometric improvement} ($S$) measures whether the rewrite
got closer to the target user's stylometric profile:
\begin{equation}
    S = d(\mathbf{v}_{\mathrm{in}},\mathbf{v}_u)
        - d(\mathbf{v}_{\mathrm{out}},\mathbf{v}_u).
\end{equation}
$d$ is cosine distance, and
$\mathbf{v}_{\mathrm{in}}$ and $\mathbf{v}_{\mathrm{out}}$ are the input and output feature
vectors. $\mathbf{v}_u$ is the mean profile formed from the user's training paragraphs. A
positive value indicates movement toward the profile. The feature vector comprises nine measures: the mean and standard deviation of sentence
and paragraph lengths, the type-token ratio, and the comma, semicolon,
em-dash, and parenthesis rates.

The per-pair composite score is
\begin{equation}
    Q = 0.40A + 0.25C + 0.15T + 0.15\widetilde{R}
        + 0.05\widetilde{S},
\end{equation}
$R$ and $S$ having different scales compared to the other metrics are normalized
to values between $[0,1]$. We define $\widetilde{R} = \frac{R+2}{4}$ and
$\widetilde{S}=\operatorname{clip}_{[0,1]}((S+0.2)/0.4)$. Under this mapping, $S=0$ corresponds to $\tilde{S}=0.5$.
The reported score for an adapter is the
mean of $Q$ over all 219 evaluation pairs. The weights of the components were fixed a priori and not
learned or validated. The sensitivity of the results to this choice
is investigated in Section~\ref{sec:automatic-results}. We report each component separately as different
combinations of content fidelity and stylistic change can lead to similar composite values.

\subsection{LLM-as-a-Judge Evaluation}
\label{sec:llm-judge-setup}
As a secondary evaluation, we measured the perceived AI-ness using five
publicly available LLM judges: ChatGPT 5.6 Sol High, Claude Sonnet 5 High,
Gemini 3.6 Flash, Grok 4.5 Expert, and Mistral Think. Ratings were collected
through the models' chat interfaces. Each judge received ten independent
batches of eight paragraphs, for 80 ratings per judge and 400 ratings in total.
A new private or temporary conversation was used for every batch.

The item pool was created from eight source paragraphs in the
held-out evaluation partition. It included four conditions: \textbf{H}
the original human paragraph; \textbf{AI}, a helper-generated
AI-shadow; \textbf{Gen}, an open-ended paragraph generated by a
helper model; and \textbf{IMS}, an InMyStyle rewrite drawn from
one of the four adapter sizes and either decoding condition. Each
batch included one item per source paragraph, ensuring that two
variants of the same source were never shown together in one
batch. Conditions and systems were hidden behind random item
keys; the mapping was not provided to the judges.

The judges provided ratings for each paragraph independently
on an integer scale from 0 (completely human-written) to 10
(completely AI-generated) and no explanations. The number of
ratings per condition across all judges was 34 H, 39 AI, 34 Gen
and 293 IMS. The per-judge condition counts varied: each judge
rated between six and eight examples from each anchor condition
and between 57 and 61 IMS outputs. The IMS assignments
covered all eight size-decoding systems, though not perfectly
balanced. We therefore report the number of observations for each
system mean. This experiment measures perceived AI-ness, not
similarity to the target user's writing.

All values in brackets are 95\% confidence intervals from a
cluster bootstrap over source paragraphs ($B=5000$): the 73
evaluation paragraphs for the automatic scores and the eight
judged paragraphs for the LLM ratings, where the intervals are
only approximate. Resampling whole paragraphs keeps all items
derived from the same paragraph together and differences
between systems are computed on the same resamples.

%% file: sections/05_results.tex
\section{Results}
\label{sec:results}

\subsection{Automatic Evaluation}
\label{sec:automatic-results}
Table~\ref{tab:automatic-results} shows all automatic metrics for the four
model sizes and two decoding conditions.
At three decimals, $Q$ ranges from 0.689 to 0.695 across the eight systems, and
all intervals overlap. Paired differences, computed on unrounded per-pair
scores, are more sensitive: 3B scores slightly below 7B under greedy
($-0.003$ $[-0.006, -0.001]$) and sampled decoding
($-0.005$ $[-0.009, -0.001]$), whereas 0.5B matches 7B under greedy
($0.000$ $[-0.004, 0.003]$). All three differences are at most 0.005 on the
0-1 scale, and the ordering does not follow model size.

\begin{table*}[t]
    \caption{Automatic evaluation on 219 pairs from 73 held-out source
    paragraphs. Higher is better for all metrics. $A$: authorship; $C$: content
    similarity; $T$: target similarity; $R$: AI-tell reduction; $S$:
    stylometric improvement; $Q$: composite. Bold marks the highest value
    within each decoding condition, including ties after rounding. Brackets give intervals.}
    \label{tab:automatic-results}
    \centering
    \small
    \setlength{\tabcolsep}{5pt}
    \begin{tabular}{llrrrrrr}
        \toprule
        Size & Decoding & $A$ & $C$ & $T$ & $R$ & $S$ & $Q$ \\
        \midrule
        0.5B & Greedy & 0.52 & \textbf{0.94} & \textbf{0.89} & \textbf{0.52} & \textbf{0.00} & 0.694 {\scriptsize[0.688, 0.701]} \\
        1.5B & Greedy & 0.52 & 0.93 & \textbf{0.89} & 0.46 & \textbf{0.00} & 0.691 {\scriptsize[0.684, 0.698]} \\
        3B   & Greedy & 0.51 & \textbf{0.94} & \textbf{0.89} & 0.47 & \textbf{0.00} & 0.691 {\scriptsize[0.685, 0.698]} \\
        7B   & Greedy & \textbf{0.53} & 0.93 & \textbf{0.89} & 0.50 & \textbf{0.00} & \textbf{0.695} {\scriptsize[0.688, 0.701]} \\
        \midrule
        0.5B & Sample & 0.52 & \textbf{0.93} & 0.88 & \textbf{0.55} & \textbf{0.00} & 0.694 {\scriptsize[0.688, 0.700]} \\
        1.5B & Sample & \textbf{0.53} & \textbf{0.93} & \textbf{0.89} & 0.53 & \textbf{0.00} & \textbf{0.695} {\scriptsize[0.689, 0.701]} \\
        3B   & Sample & 0.52 & 0.92 & \textbf{0.89} & 0.45 & \textbf{0.00} & 0.689 {\scriptsize[0.682, 0.697]} \\
        7B   & Sample & \textbf{0.53} & 0.92 & \textbf{0.89} & 0.49 & \textbf{0.00} & 0.694 {\scriptsize[0.687, 0.701]} \\
        \bottomrule
    \end{tabular}
\end{table*}

Authorship stays close to the classifier's decision boundary for all systems.
The measured score is dominated by high semantic fidelity, moderate
reduction in AI-tell and small movements in the two explicit style metrics. Sampling does not seem to have a consistent effect on the composite. Across
sizes, it slightly lowers content similarity and increases authorship. Its effect
on AI-tell reduction is mixed: it improves the metric for 0.5B and 1.5B but
reduces it for 3B and 7B. Overall, the two decoding conditions are comparable, 
and we suggest greedy decoding because it is deterministic.

The stylometric improvement $S$ remains close to zero (between $-0.004$ and $0.002$) 
for all the systems. Two properties explain this. First, the nine features are not
standardized before the cosine distance, so paragraph length (about 290
words on average) dominates the vector, while punctuation rates (below
0.1) barely register. Second, the rewrites keep paragraph length close to
the input (median output-to-input length ratio of 0.96-1.01 across systems),
so the dominant feature hardly changes. The measure also does not capture
the lexical changes, such as removing formulaic framing, that the adapters
mainly make. Because $\tilde{S}\approx 0.5$ for every system, $S$ adds a
near-constant $0.05 \times 0.5 = 0.025$ to $Q$ and does not
affect the comparison. Together with authorship scores near 0.5, this indicates that the adapters mainly 
naturalize the text rather than recover a distinctive personal style.

To test the sensitivity to the component weights, we re-computed $Q$ 
from the per-pair component scores with equal weights, with $S$
removed and the remaining weights renormalized, and with each weight
in turn halved or increased by 50\%. The level of $Q$ moves between
0.66 and 0.74 across these schemes, but the spread across the eight systems
stays at or below 0.008. No scheme monotonically orders the greedy
systems by model size. Thus, the value 0.69 is sensitive to the chosen
weights, but the flatness across model sizes is not.

Training cost provides a distinction among the models. As
shown in Table~\ref{tab:training-dynamics}, the wall time
increases from 9.53 minutes for 0.5B to 67.73 minutes for 7B
on the same hardware and recipe. The mean training crossentropy (CE) decreases with the model size. Moreover, the
larger models also achieve lower evaluation cross-entropy. However, this ordering does not carry over to the task level composite.

\begin{table}[t]
    \caption{Training dynamics by model size. Train CE is the mean training
    loss recorded over three epochs; Eval CE$_1$ and Eval CE$_3$ are evaluation
    losses after the first and third epochs.}
    \label{tab:training-dynamics}
    \centering
    \small
    \begin{tabular}{lrrrr}
        \toprule
        Size & Train CE & Eval CE$_1$ & Eval CE$_3$ & Minutes \\
        \midrule
        0.5B & 1.45 & 1.66 & 1.66 &  9.53 \\
        1.5B & 1.26 & 1.47 & 1.51 & 20.10 \\
        3B   & 1.18 & 1.40 & 1.46 & 35.67 \\
        7B   & 1.05 & 1.33 & 1.43 & 67.73 \\
        \bottomrule
    \end{tabular}
\end{table}

\subsection{Perceived AI-ness}
\label{sec:judge-results}
Table~\ref{tab:judge-condition-results} summarizes the ratings on
the 0-10 scale, where lower values indicate more human-like text. Pooled across judges, human paragraphs receive the lowest mean
score ($3.47$), followed by InMyStyle rewrites ($5.34$), helper AI shadows ($6.85$), and open-ended generated text ($8.44$). The mean
perceived-AI score of InMyStyle rewrites is $22\%$ lower than that of
their helper-generated inputs. The separation of Gen suggests that
most judges readily identify open-ended generated prose, while the
paired H, AI and IMS conditions are harder to
distinguish.

\begin{table}[t]
    \caption{Perceived AI-ness by judge and condition. Lower is more
    human-like. Counts: H 34, AI 39, IMS 293, and Gen 34.}
    \label{tab:judge-condition-results}
    \centering
    \small
    \begin{tabular}{lrrrr}
        \toprule
        Judge & H & AI & IMS & Gen \\
        \midrule
        ChatGPT               & 1.50 & 7.88 & 4.83 & 9.25 \\
        Claude                & 2.29 & 5.50 & 4.86 & 9.14 \\
        Gemini                & 1.29 & 5.50 & 3.88 & 8.67 \\
        Grok                  & 4.75 & 7.50 & 5.82 & 8.71 \\
        Mistral               & 7.67 & 8.00 & 7.23 & 6.00 \\
        \midrule
        All judges            & 3.47 & 6.85 & 5.34 & 8.44 \\
        \bottomrule
    \end{tabular}
\end{table}

The per-judge rows show a fair amount of calibration
discrepancy. ChatGPT and Gemini are at the lower end of the spectrum
for H and IMS compared to AI and Gen. Mistral on the other hand
squeezes all four conditions into the $6.00$-$8.00$ range and rates Gen as
less AI-like than the other three. Hence the pooled means should be
understood as a multi-model proxy and not as interchangeable or
equivalent human calibrated judgments.

In the IMS conditions, the pooled ratings tend to get more
human-like with the increase of model size (Table~\ref{tab:judge-system-results}). The size-pooled means are
$6.24$, $5.80$, $4.93$ and $4.45$ for 0.5B, 1.5B, 3B and 7B
respectively. The intervals of adjacent sizes overlap. 0.5B and 1.5B are rated as more AI-like than 7B 
(differences 1.78
$[0.55, 3.32]$ and 1.35 $[0.36, 2.50]$), whereas the 3B--7B difference
(0.48 $[-0.29, 1.38]$) lies within the uncertainty. The size
ordering is therefore suggestive rather than conclusive.

\begin{table}[t]
    \caption{Perceived AI-ness of IMS outputs by size and decoding; lower is
    more human-like. Brackets give intervals.}
    \label{tab:judge-system-results}
    \centering
    \small
    \begin{tabular}{lrr|rr|rr}
        \toprule
        & \multicolumn{2}{c}{Greedy} & \multicolumn{2}{c}{Sample}
        & \multicolumn{2}{c}{Pooled} \\
        Size & $n$ & Mean & $n$ & Mean & $n$ & Mean \\
        \midrule
        0.5B & 36 & 6.39 & 36 & 6.08 & 72 & 6.24 {\scriptsize[5.27, 7.24]} \\
        1.5B & 36 & 5.39 & 35 & 6.23 & 71 & 5.80 {\scriptsize[4.90, 6.67]} \\
        3B   & 38 & 4.95 & 35 & 4.91 & 73 & 4.93 {\scriptsize[3.56, 6.31]} \\
        7B   & 37 & 4.11 & 40 & 4.78 & 77 & 4.45 {\scriptsize[3.47, 5.33]} \\
        \bottomrule
    \end{tabular}
\end{table}

\subsection{Qualitative De-AI Diagnostics}
\label{sec:qualitative-results}
To further illustrate the compact models, we evaluate them on three general-domain AI-generated passages rich with AI associated wording. The
chosen passages were constructed to
contain formulaic framing and are not representative of the
target user's prose. Thus, they are presented to demonstrate
naturalization rather than authorship recovery. Table~\ref{tab:qual-metrics} shows the automatic diagnostics of the passages, and
Appendix~\ref{sec:qualitative-appendix} presents selected transformations.

\begin{table}[t]
    \caption{Automatic diagnostics averaged over three general-domain
    demonstrations. Content $F_1$ is BERTScore against the AI input; AI-tell
    reduction uses the same phrase/substitution dictionary as the primary automatic evaluation. Tells-out is the mean number of configured AI-tell occurrences remaining in the text. 
    Target \(F_1\), authorship and stylometric improvement are not reported, as these passages have no user-written reference. Among IMS rows, bold marks the best value in each column, including ties after rounding.}
    \label{tab:qual-metrics}
    \centering
    \small
    \begin{tabular}{lrrr}
        \toprule
        System & Content $F_1$ & AI-tell reduction & Tells out \\
        \midrule
        AI input & 1.00 & 0.00 & 11.7 \\
        \rowcolor{imsrow}
        IMS 0.5B greedy & \textbf{0.94} & 0.76 & 2.7 \\
        \rowcolor{imsrow}
        IMS 1.5B greedy & \textbf{0.94} & 0.88 & 1.3 \\
        \rowcolor{imsrow}
        IMS 3B greedy & \textbf{0.94} & 0.83 & 2.0 \\
        \rowcolor{imsrow}
        IMS 7B greedy & 0.93 & \textbf{0.91} & \textbf{1.0} \\
        \bottomrule
    \end{tabular}
\end{table}

The AI-input row represents the baseline. All four
IMS models achieve high mean content similarity ($0.93$-$0.94$) and reduce the number of configured
tells. The 7B model achieves the largest mean AI-tell
reduction ($0.91$) and the fewest tells out ($1.0$). The intermediate models are between the two endpoint systems
on these diagnostics. At the sample level, 0.5B
under-edits Sample~A and thins some content in Sample~B. In Sample~C, 7B does not dominate every individual
example.

\subsection{Automatic and Perceived-AI Assessment}
\label{sec:ranking-disagreement}
All the models remain viable under the primary composite, while training cost and the secondary perceived-AI ratings distinguish their practical choice of use. 
The 3B model offers a practical sweet spot with a composite score of 0.691, within 0.004 of 7B, and greedy content similarity of 0.94 and a pooled
perceived-AI score of 4.93, indistinguishable from 7B, but with
47\% less training time. The smaller 0.5B and 1.5B models remain suitable when 
compute is more constrained, while 7B provides the lowest perceived AI-ness when compute is a 
secondary concern.

%% file: sections/06_limitations.tex
\section{Limitations}
\label{sec:limitations}
The experiments use documents from one target user and therefore
cannot establish generalization across users, languages or amounts of
available writing. The publication timespan and co-authorships cause the
learned distribution to include temporal, domain, co-author, reviewer and
venue effects rather than an uncontaminated personal writing style. The training inputs are also synthetic: three helper-model families and a
manually configured corruption process approximate, but cannot cover,
the variety of AI-edited or generated prose encountered in practice. Finally, the model-size sweep is restricted to the Qwen2.5 family and one
rank-8, three-epoch recipe; it does not establish that the same plateau
holds for other model families or adaptation settings.

The automatic results are specific to a manually weighted composite,
although the flatness across model sizes is robust to re-weighting
(Section~\ref{sec:automatic-results}). The authorship classifier may capture general
human-machine or helper-specific signals in addition to user
style, BERTScore can reward conservative copying, and the
stylometric profile captures only coarse surface features.

The study also lacks two reference points: an unadapted (no-LoRA)
base model and a prompt-only personalization baseline, e.g., few-shot
prompting with the user's paragraphs as in-context examples. Future work will 
quantify how much of the naturalization effect is due
to per-user fine-tuning as opposed to the base model or to prompting
alone.

The LLM-judge study contains 400 ratings but is based on eight source
paragraphs and uses unequal sample counts per IMS
system. The judges are calibrated very differently to each
other, are external models that may change over time, and
measure perceived AI-ness rather than similarity to the target
user. No completed human-participant ratings are available.

%% file: sections/07_conclusion.tex
\section{Conclusion}
\label{sec:conclusion}
InMyStyle is a local, single-user system that learns to rewrite AI-edited
paragraphs toward the language distribution of a user. Paired supervision is
constructed by transforming user-written paragraphs with multiple local
helper-model families. Source leakage is prevented by paragraph-level
splitting. Compact user-specific LoRA adapters are trained with response-only
loss. Length-aware generation budget and sentence-boundary chunking
provide a common inference path for interactive rewriting and automatic
evaluation.

Across the Qwen2.5 models with
parameters ranging from 0.5B to 7B, the defined automatic composite score varies between 0.689
and 0.695 for both greedy and sampled decoding. Thus, in
this experiment, the larger model size does not have a
consistent advantage on the composite score. Relatively small models also achieve 
similar content fidelity to the 7B model with much lower training cost. 

The secondary LLM-judge study showed that InMyStyle outputs were
perceived as less AI-like than their helper-generated inputs. Within InMyStyle, 
the mean perceived-AI scores tended to decrease with model size, though the calibration varied substantially across judges. We conclude that automatic task metrics
and perceived-AI judgments should be seen as complementary
but not interchangeable pieces of evidence.

For the dataset, model family, training recipe and composite metric
evaluated in this paper, the results support the view that small can be
enough for the measured rewriting task. It supports content-preserving naturalization more strongly than
recovery of personal style. Based on the evaluation results, 
the 3B model balances performance and efficiency better, though users prioritizing lower perceived AI-ness 
over compute cost may still favor the 7B variant. Establishing broader per-user style transfer remains a 
target for future work, which will incorporate expanded user corpora, base-model and prompt-based baselines, 
and blinded human evaluation.

%% file: sections/08_qualitative_appendix.tex
\section{Appendix: Qualitative De-AI Examples}
\label{sec:qualitative-appendix}
Tables~\ref{tab:qual-sample-a}-\ref{tab:qual-sample-c} show excerpts from
the three general-domain demonstrations. Inputs deliberately uses
formulaic expressions such as \textit{furthermore}, \textit{in essence}
and \textit{utilize} to make naturalization visible. The 0.5B outputs vary the most, under-editing
Sample~A and thinning Sample~B. Tell removal is not ordered by size: on
average, 1.5B leaves fewer tells than 3B. The 7B outputs remove stacked
discourse frames most consistently, although one configured tell remains
in Sample~C.

\begin{table*}[t]
    \caption{Sample A (pangram and typography).
    \aitell{Red} marks AI-associated framing or wording,
    \changed{green} marks clearer naturalization, and
    \partialfix{yellow} marks partial cleanup or content thinning. Ellipses mark omitted text, not model deletions beyond those
    explicitly noted.}
    \label{tab:qual-sample-a}
    \centering
    \footnotesize
    \renewcommand{\arraystretch}{1.2}
    \begin{tabularx}{\textwidth}{@{}>{\raggedright\arraybackslash}p{1.7cm} X@{}}
        \toprule
        \textbf{System} & \textbf{Selected text} \\
        \midrule
        AI input
        & \aitell{Furthermore,} the quick brown fox is commonly
        \aitell{utilized} as an illustrative example \ldots\
        \aitell{In essence,} this construction is not intended to describe a
        real wildlife observation; rather, it is \aitell{utilized} to
        \aitell{facilitate} typographic and pedagogical tasks.
        \aitell{Moreover,} educators and designers frequently rely on the
        sentence \ldots\ \aitell{It is worth mentioning that} alternative
        pangrams exist \ldots\ \aitell{due to the fact that} it balances
        simplicity with coverage. \\
        \rowcolor{imsrow}
        IMS 0.5B
        & \aitell{Furthermore,} the quick brown fox is commonly
        \changed{used} as an illustrative example \ldots\
        \aitell{In essence,} this construction \ldots\ is
        \aitell{utilized} to \aitell{facilitate} typographic and pedagogical
        tasks. \aitell{Moreover,} educators and designers \ldots\
        \aitell{It is worth mentioning that} alternative pangrams exist
        \ldots\ \aitell{due to the fact that} it balances simplicity with
        coverage. \partialfix{The final two input sentences are omitted.} \\
        \rowcolor{imsrow}
        IMS 1.5B
        & \aitell{Moreover,} the quick brown fox is often
        \changed{used} as an example when discussing character coverage
        \ldots\ This construction is not intended to represent a real
        wildlife observation, but is used to \aitell{facilitate} typographic
        and pedagogical tasks \ldots\ The example below continues to \aitell{demonstrate} \ldots\
\partialfix{Three configured tells remain.} \\
        \rowcolor{imsrow}
        IMS 3B
        & \changed{The quick brown fox jumps over the lazy dog} is a common
        pangram used to \aitell{demonstrate} character coverage \ldots\
        The construction is not meant to describe a real wildlife
        observation, but rather to \aitell{facilitate} typographic and
        pedagogical tasks \ldots\ The example continues to \aitell{demonstrate} \ldots\
\partialfix{Three configured tells remain.} \\
        \rowcolor{imsrow}
        IMS 7B
        & \changed{The quick brown fox} jumps over the lazy dog is a well-known
        pangram, a sentence that incorporates every letter of the alphabet.
        It is often \changed{used} as an example \ldots\ The sentence is not
        intended to describe a real observation of wildlife, but is used to
        \aitell{facilitate} typographic and pedagogical tasks \ldots\
        \changed{Opening frames removed}; most content is retained. \\
        \bottomrule
    \end{tabularx}
\end{table*}

\begin{table*}[t]
    \caption{Sample B (morning coffee). Colors and ellipses follow
    Table~\ref{tab:qual-sample-a}.}
    \label{tab:qual-sample-b}
    \centering
    \footnotesize
    \renewcommand{\arraystretch}{1.2}
    \begin{tabularx}{\textwidth}{@{}>{\raggedright\arraybackslash}p{1.7cm} X@{}}
        \toprule
        \textbf{System} & \textbf{Selected text} \\
        \midrule
        AI input
        & Many individuals \aitell{utilize} a morning cup of coffee in order
        to \aitell{commence} their daily routine. \aitell{Indeed,} the
        beverage is frequently associated with increased alertness \ldots\
        \aitell{In essence,} preparing coffee can be viewed as both a
        practical task and a small personal ritual. \aitell{Moreover,} some
        prefer a simple black brew \ldots\ \aitell{It is important to note
        that} grinding beans immediately before brewing is often recommended
        to \aitell{facilitate} a fresher aroma. \\
        \rowcolor{imsrow}
        IMS 0.5B
        & \changed{Many people prefer to drink coffee in the morning.} Coffee
        is often associated with increased alertness \ldots\ Some people
        prefer a strong coffee, while others prefer a weak one \ldots\
        \aitell{It is important to note that} grinding the coffee beans right
        before brewing can help produce a fresher aroma.
        \partialfix{The brewing-method and social-custom content is dropped.}
        \\
        \rowcolor{imsrow}
        IMS 1.5B
        & \changed{Coffee is a beverage that is often consumed in the morning to kick start} the day. It is generally associated with increased
        alertness, although the effect can vary with tolerance and sleep
        quality \ldots\ Cafés and home kitchens vary in their methods of
        brewing coffee, from pour-over to espresso machines \ldots\
        \changed{All configured tells are removed and the main content is
        retained.} \\
        \rowcolor{imsrow}
        IMS 3B
        & \changed{Coffee is a beverage that many people drink in the morning
        in order to} start their day. \ldots\ Some prefer a simple black brew,
        while others add milk, sugar, or flavored syrups to
        \aitell{obtain} a sweeter profile \ldots\ Cafés and home kitchens
        alike \aitell{demonstrate} a wide range of methods \ldots\
        \partialfix{Two configured tells remain; the main content is
        retained.} \\
        \rowcolor{imsrow}
        IMS 7B
        & \changed{Many people start their day} with a cup of coffee. It is
        often associated with increased alertness, although the effect varies
        with caffeine tolerance and sleep \ldots\ Making coffee is both a
        practical task and a small personal ritual \ldots\ Different brewing
        methods result in different body, bitterness, and serving size.
        \changed{Most frames removed}; the main content is retained. \\
        \bottomrule
    \end{tabularx}
\end{table*}

\begin{table*}[t]
    \caption{Sample C (city park). Colors and ellipses follow
    Table~\ref{tab:qual-sample-a}.}
    \label{tab:qual-sample-c}
    \centering
    \footnotesize
    \renewcommand{\arraystretch}{1.2}
    \begin{tabularx}{\textwidth}{@{}>{\raggedright\arraybackslash}p{1.7cm} X@{}}
        \toprule
        \textbf{System} & \textbf{Selected text} \\
        \midrule
        AI input
        & \aitell{In essence,} a city park can be \aitell{utilized} by
        residents seeking open space within a dense urban environment
        \ldots\ \aitell{Furthermore,} the park functions as a shared public
        resource. \aitell{Moreover,} on warm afternoons one may observe
        joggers, families, and readers \ldots\ \aitell{Notably,} local
        regulations may limit amplified music or commercial activity
        \ldots\ \aitell{In conclusion,} the park illustrates how simple
        outdoor amenities can support community well-being. \\
        \rowcolor{imsrow}
        IMS 0.5B
        & \changed{A city park can be a place} for people to have open space
        in a dense urban environment. The walking paths, benches, and shaded
        lawns provide a place for casual recreation \ldots\ The park is a
        shared public resource, not a private garden \ldots\
        \partialfix{It is worth noting that} seasonal plantings and water
        features can encourage longer visits.
        \changed{Most heavy discourse frames are removed.} \\
        \rowcolor{imsrow}
        IMS 1.5B
        & \changed{A city park can be a place for residents to enjoy open
        space in a dense} urban environment. Walking paths, benches, and
        shaded lawns provide opportunities for casual recreation \ldots\
        Evening hours can \aitell{demonstrate} a quieter pattern of use as
        daylight decreases.
        \partialfix{One configured tell remains; the final
        community sentence is omitted.} \\
        \rowcolor{imsrow}
        IMS 3B
        & \changed{A city park can provide residents with a green space in
        the midst of a} dense urban environment. Walking paths, benches, and
        shaded lawns provide opportunities for casual recreation \ldots\
        Evening hours can \aitell{demonstrate} a quieter pattern of use as
        daylight decreases. The park illustrates how simple outdoor amenities
        can support community well-being without requiring complex
        infrastructure.
        \partialfix{One configured tell remains; the core content is
        retained.} \\
        \rowcolor{imsrow}
        IMS 7B
        & \changed{A city park is a place} where people can go to get some open
        space in the middle of a dense urban environment. There are walking
        paths, benches, and shaded lawns that \aitell{facilitate} casual
        recreation \ldots\ It is a shared public resource rather than a
        private garden \ldots\ \changed{In the evening,} the pattern of use is
        quieter as daylight diminishes.
        \partialfix{Most frames are removed, with one configured tell
        remaining.} \\
        \bottomrule
    \end{tabularx}
\end{table*}